\documentclass[letterpaper, 10 pt, conference]{ieeeconf}
\IEEEoverridecommandlockouts 
\usepackage{hyperref}
\usepackage{graphicx}
\usepackage{amsmath,amssymb}
\usepackage{cite}
\usepackage{float}
\usepackage[caption=false,font=footnotesize]{subfig}

\renewcommand{\vec}{\boldsymbol}

\graphicspath{{img/},{fig/}}

\begin{document}

\title{\LARGE \bf
A Fast Gear-Shifting Actuator for Robotic Tasks with Contacts
}

\author{Alexandre Girard$^{1}$ and H. Harry Asada$^{2}$
\thanks{$^{1}$Alexandre Girard is with the Department of Mechanical Engineering, Universite de Sherbrooke, Qc, Canada {\tt\small  alex.girard@usherbrooke.ca }}%
\thanks{$^{2}$H. Harry Asada is with the Department of Mechanical Engineering, Massachusetts Institute of Technology, Cambridge, MA, USA {\tt\small asada@mit.edu } }
}%

\maketitle
\thispagestyle{empty}
\pagestyle{empty}

\begin{abstract}
Vehicle power-trains use a variable transmission (multiple gear-ratios) to minimize motor size and maximize efficiency while meeting a wide-range of operating points. Robots could similarly benefit from variable transmission to save weight and improve energy efficiency; leading to potentially groundbreaking improvements for mobile and wearable robotic systems. However, variable transmissions in a robotic context leads to new challenges regarding the gear-shifting methodology: 1) order-of-magnitude variations of reduction ratios are desired, and 2) contact situations during manipulation/locomotion tasks lead to impulsive behavior at the moment when gear-shifting is required. This paper present an actuator with a gear-shifting methodology that can seamlessly change between two very different reduction ratios during dynamic contact situations. Experimental results demonstrate the ability to execute a gear-shift from a 1:23 reduction to a 1:474 reduction in less than 30ms during contact with a rigid object.
\end{abstract}

\section{Introduction}

\begin{figure}[h]
 \vspace{-5pt}
	\centering
		\includegraphics[width=0.45\textwidth]{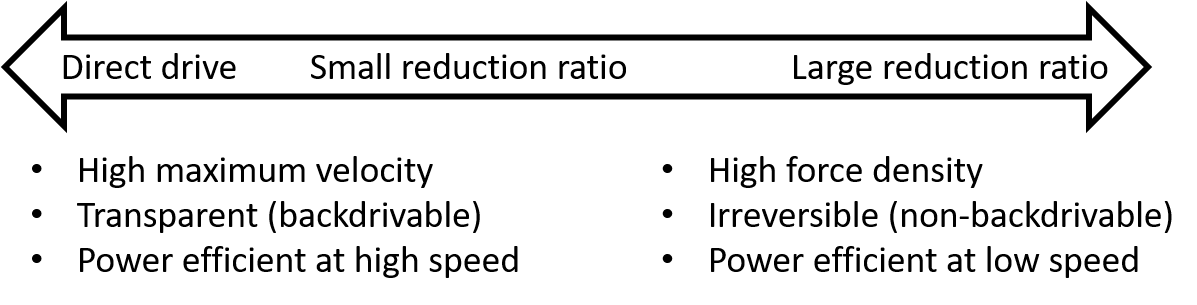}
		\vspace{-10pt}
	\caption{Example of major trade-offs when selecting a robot reduction ratios}
	\vspace{-5pt}
	\label{fig:tradeoff}
\end{figure}

The selection of actuator reduction ratios is highly critical for robotic systems. It influences payload capability, energy efficiency, maximum speed, and also a robot's intrinsic impedance. With single-gear electric motors, a robot designer has to heavily compromise, as shown in Fig. \ref{fig:tradeoff}.

Variable transmission actuators that can dynamically change their reduction ratio, like in car power-trains, can enable fast and strong lightweight robots. Even a small, lightweight electric motor can generate large forces and move at high speed if equipped with both large and small gear ratios. Moreover, variable transmission can enable dynamic adjustments of the intrinsic impedance of the robot joints to meet specific task requirements, such as 1) back-drivability with small reduction ratios for safe interactions with the environment or 2) irreversibility with very large reduction ratios to support large payloads without consuming energy.

\begin{figure}[t]
	\centering
		\includegraphics[width=0.40\textwidth]{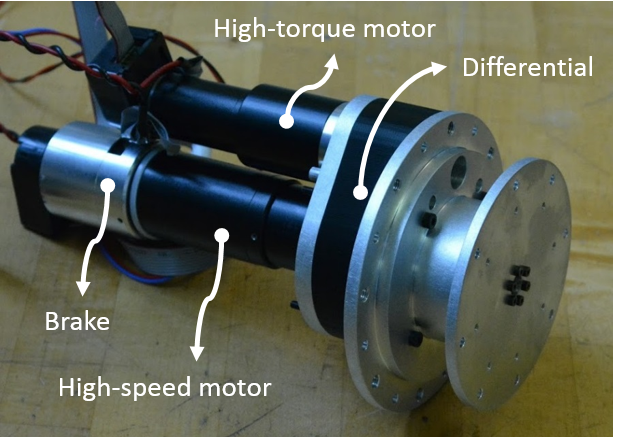}
		\vspace{-10pt}
	\caption{Dual-speed dual-motor (DSDM) actuator prototype with two possible effective reduction ratios of 1:23 and 1:474}
	\vspace{-5pt}
	\label{fig:dsdm_proto}
\end{figure}

\begin{figure}[t]
	\centering
		\includegraphics[width=0.47\textwidth]{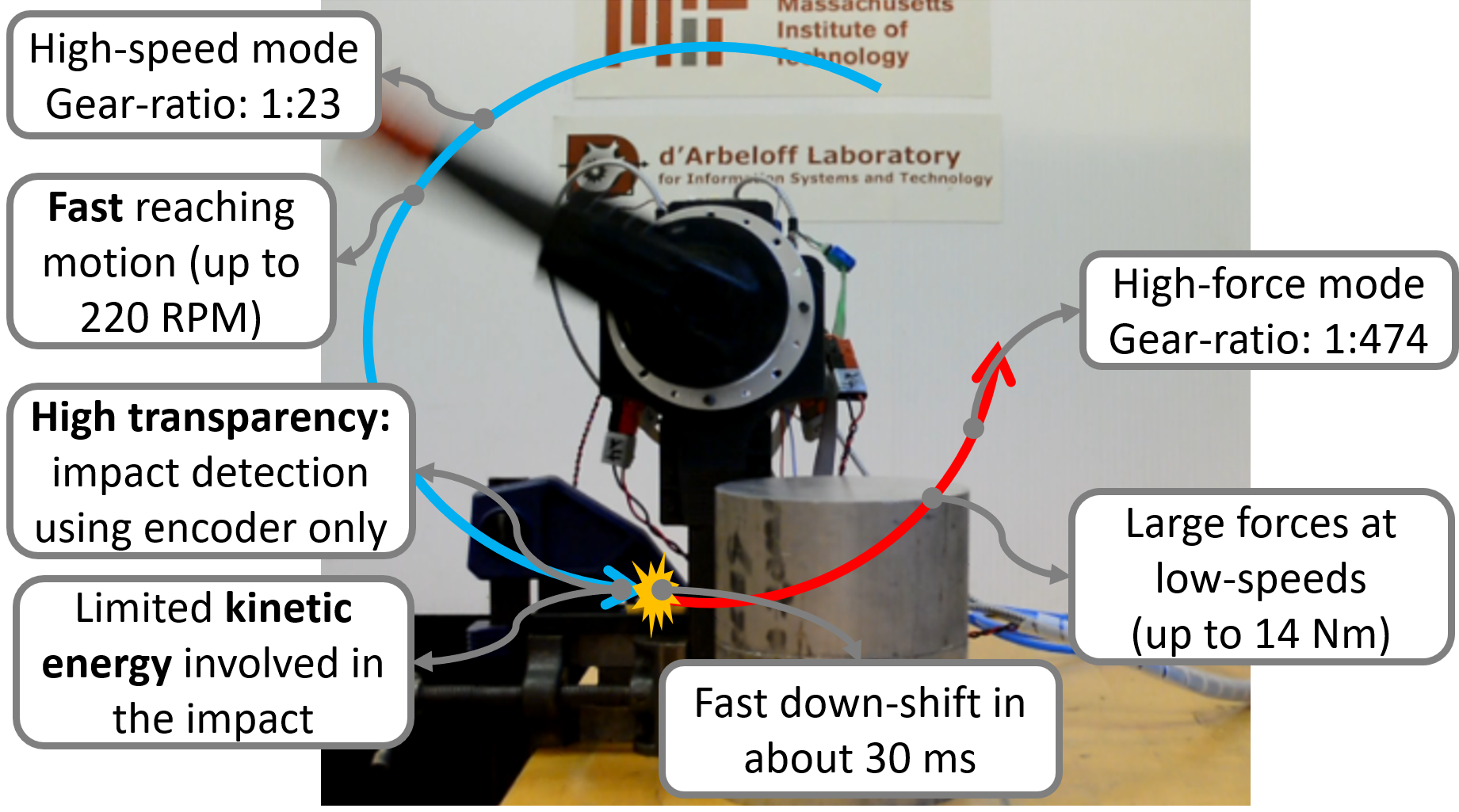}
		\vspace{-10pt}
	\caption{Illustration of the actuator capabilities in the context of a reaching and pushing task leveraging the two reduction ratios, the video of the experiment is available at \url{https://youtu.be/Aq-IQ_sZRRI}}
	\vspace{-15pt}
	\label{fig:dsdm_demo}
\end{figure}

\begin{figure*}[t]
        \centering
        \subfloat[Locomotion]{
				\includegraphics[width=0.25\textwidth]{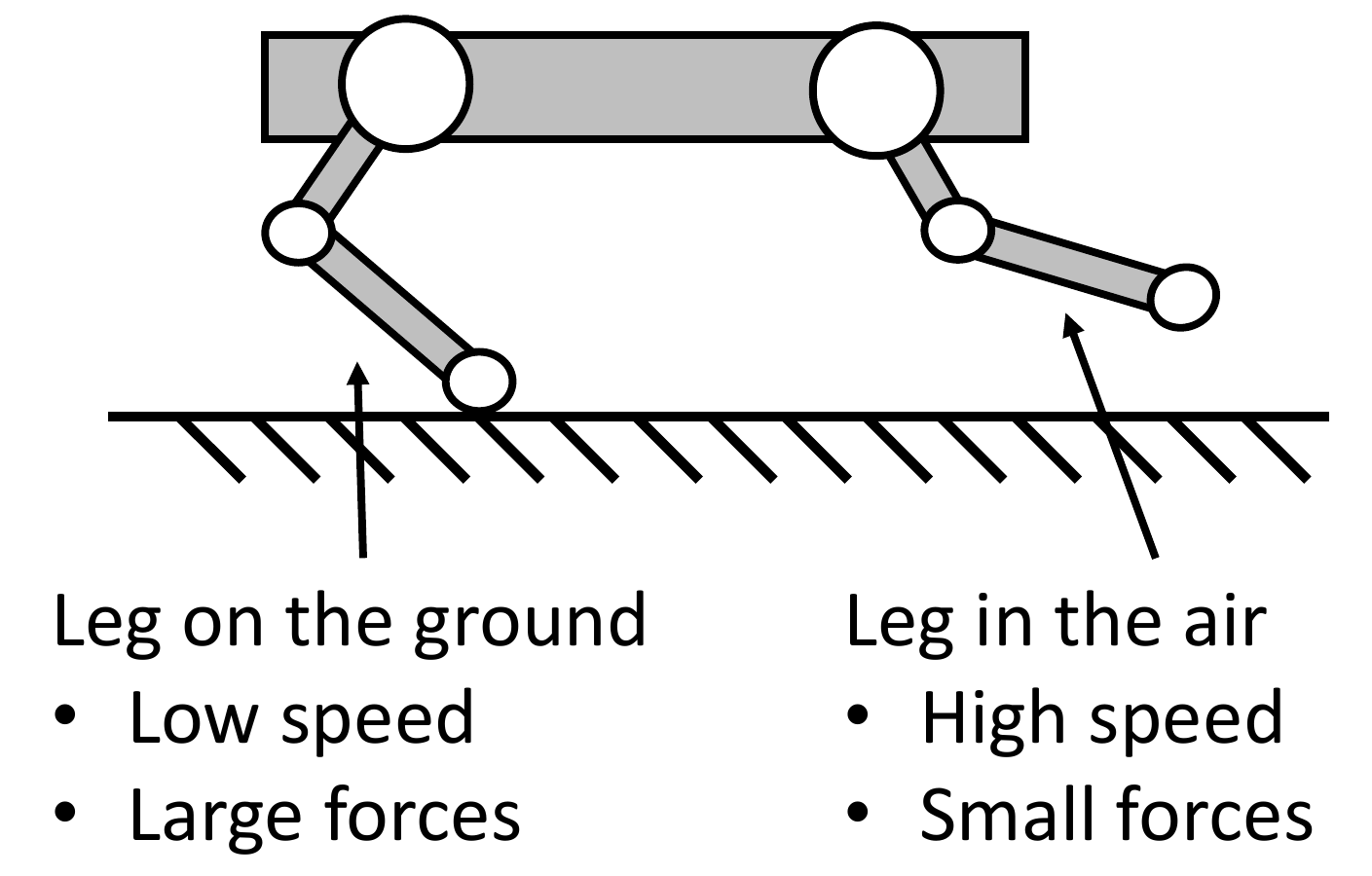}
				\label{fig:leggedrobot}}
				\hspace{15pt}
				\subfloat[Manipulation]{
				\includegraphics[width=0.23\textwidth]{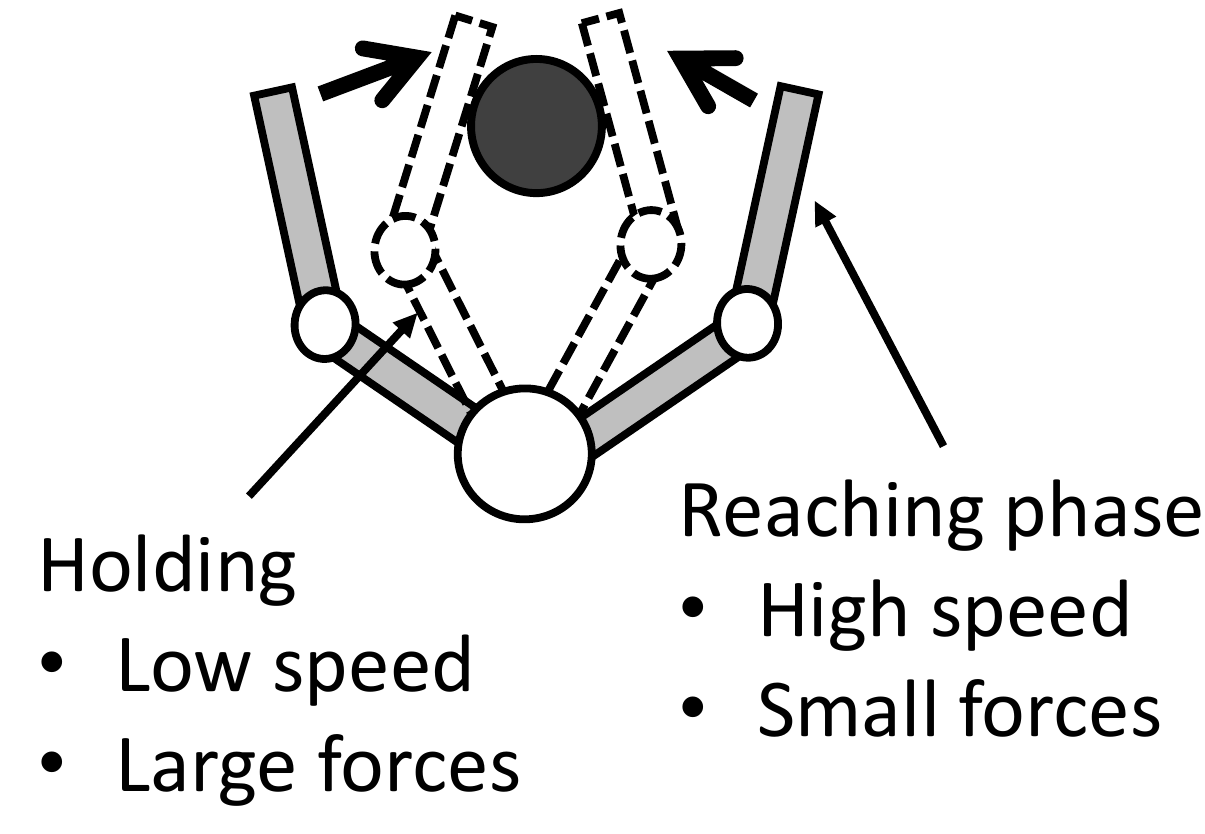}
				\label{fig:gripper}}
				\hspace{15pt}
				\subfloat[Machining]{
				\includegraphics[width=0.23\textwidth]{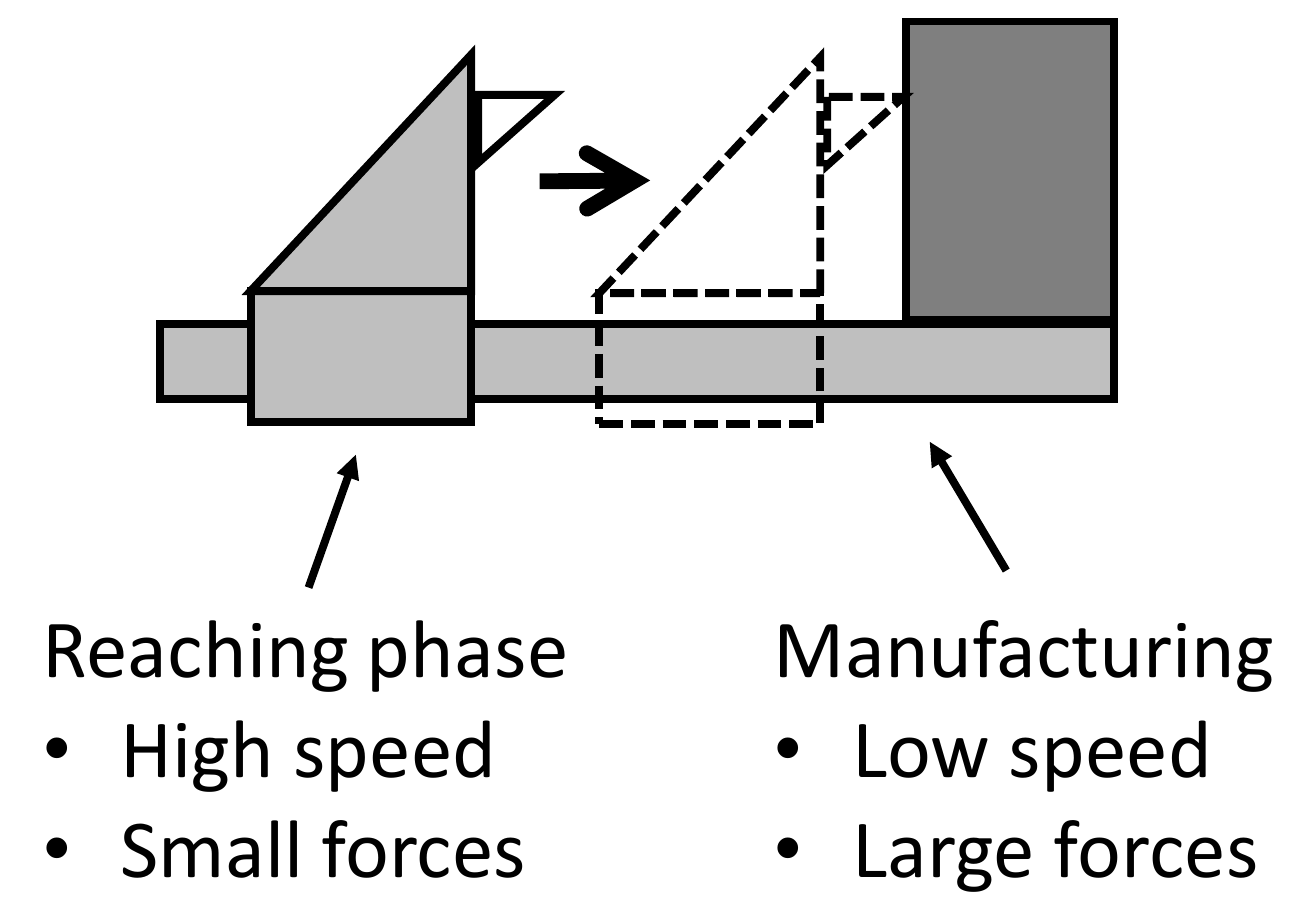}
				\label{fig:machinetool}}
        \caption{Bi-modal tasks where actuators must rapidly change from high-speed/low-force to high-force/low-speed after making contact with a load}
				\label{fig:app}
				\vspace{-12pt}
\end{figure*}

This paper discusses an actuator technology, called DSDM, with two discrete reduction ratios and the capability to change between the two very quickly in contact situations. Fig. \ref{fig:dsdm_proto} shows an actuator prototype and Fig. \ref{fig:dsdm_demo} highlights the desirable features of the technology in the context of a manipulation experiment available in the video attachment. As opposed to gear-shifting technologies typically used by vehicle power-trains, the proposed approach enable seamless transitions between order-of-magnitude different reduction ratios during highly dynamic situations such as contact impacts. As illustrated in Fig. \ref{fig:app}, this variable transmission technology addresses the needs of many robotic tasks, such as manipulation and locomotion. In the following, section \ref{sec:rel} discusses related works and limitations of state-of-the-art actuation technologies, section \ref{sec:DSDM} presents the DSDM actuator, section \ref{sec:mod} presents dynamic models and control algorithms for coordinating motors for fast transitions during impacts, and section \ref{sec:exp} presents experimental results.

\section{Background}
\label{sec:rel}
\begin{figure}[htb]
				\vspace{-20pt}
        \centering
				\subfloat[Electromagnetic transducer]{
				\includegraphics[width=0.20\textwidth]{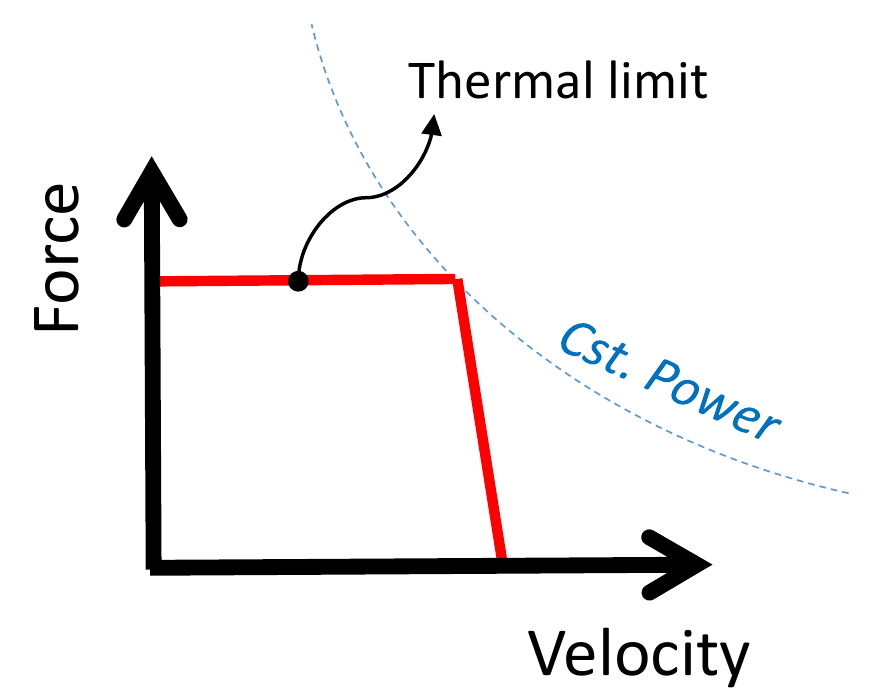}
				\label{fig:emcurve}
				}
        \subfloat[Fluidic transducer]{
				\includegraphics[width=0.18\textwidth]{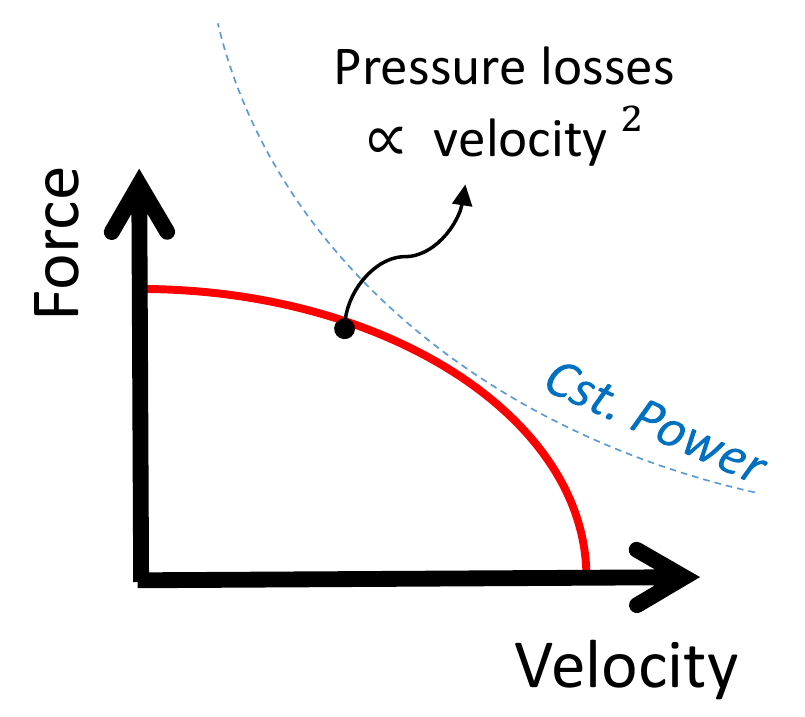}
				\label{fig:fluidcurve}
				}
        \caption{Typical force-speed curve of actuators. Since the power output is not constant for all speed, a variable transmission can be advantageous}\label{fig:powercurves}
				\vspace{-5pt}
\end{figure}

In terms of power-throughput, as illustrated in Fig. \ref{fig:emcurve}, electric actuators are typically characterized by a flat force curve for most of their range of operating speed, leading to maximum power only being available at high velocity. Some advanced electric motor systems can extend their operation at high-speed by weakening the magnetic flux, and can thus transmit their maximum power over a wider range of speeds. However, this scheme cannot go around the fundamental force saturation at low-speed, which is limited by material properties \cite{hollerbach_comparative_1992}. Fluidic actuators are typically characterized by force curves dropping quadratically with velocity (related to pressure losses in valves orifices), as illustrated in Fig. \ref{fig:fluidcurve}. All in all, EM and fluidic actuators are not perfect power sources and could benefit from using variable transmissions to have their maximum power available on a much wider range of speeds.

In the vehicle power-train field, the torque-speed matching issue with variable transmission is predominant since power density and efficiency are critical for mobile systems. Internal combustion engine have a very narrow peak of efficient power transmission and thus must absolutely be equipped with variable transmission. For electric motorizations, the idea of using multiple gear ratios to improve efficiency and power density has also been explored \cite{bologna_electric_2014} \cite{lacerte_design_2016}. While power-trains type of variable transmission provide effective solutions for torque-speed matching, they are not adapted to the robotic context. First, because power-train gear-shifting methodology are not adapted to make shifts in highly dynamic situation like contacts. Second, because they are designed to make shifts between gear-ratios very close one to another. Third, because from a design perspective, advance power-train systems are very complex machines (wet clutches, dry clutches, synchronizers, etc.), leading to manufacturing, maintenance, wear and reliability challenges. Adding such systems, in all the many actuators of a robotic system might be especially hard to justify when balancing all those practical issues.


Continuously variable transmissions (CVT) have the advantage of avoiding the gear-shifting issues. Most common designs are based on belts with variable-diameter pulleys or toroidal disks. Typical CVT have a limited total ratio variation range (typically 2x-3x), which makes them un-adapted for very large ratio variations. Some designs have been proposed for infinite range variation, often called IVT \cite{schoolcraft_gear_2011}\cite{kembaum_ultra-compact_2017}. However, since those designs rely on friction, torque-density is highly limited, which inhibits potential use for most robotic applications. Also, lever mechanisms have been proposed with variable attachment points that lead to a very wide range of effective transmission ratio \cite{tahara_high-backdrivable_2011}, however this drastically limits the output motion range. All in all, many clever mechanisms have been proposed to be used as CVT. However, despite many decades of development, all designs have major drawbacks regarding either: total variation range, constraints on output motion, ratio-variation speed, allowable shift conditions, efficiency, etc.

While variable transmissions have been studied extensively for automobile power-trains, they have not yet been fully investigated in robotics despite significant potential gains. A few instances of research in that direction were made for legged locomotion \cite{hirose_design_1991}, grasping robotic hands with twisting cable mechanisms \cite{jeong_design_2017}, actuators using ball-screws \cite{hirose_development_1999}, planetary differentials \cite{kim_improved_2007} \cite{lee_new_2012} \cite{girard_two-speed_2015}\cite{verstraten_modeling_2018}, and gear-clutches \cite{phlernjai_jam-free_2017}. While these works are promising, none have studied dynamic mode transitions during impacts and proposed adapted fast gear-shifting methodology, which is a potentially enabling feature for many robotic applications. 
%
%
%
This paper build on the DSDM actuation concept \cite{girard_two-speed_2015} and tackle the issue of contact situations, which is critical for locomotion and manipulation tasks. The main novel contribution is the methodology and demonstration of gear shifting between two order-of-magnitude different gear-ratios seamlessly in contact situations. Furthermore, this paper presents a novel compact actuator prototype, an analysis of the hybrid behavior and adapted control algorithms. 



\section{Dual-Speed Dual-Motor Actuators}
\label{sec:DSDM}

\begin{figure}[htbp]
	\centering
		\includegraphics[width=0.49\textwidth]{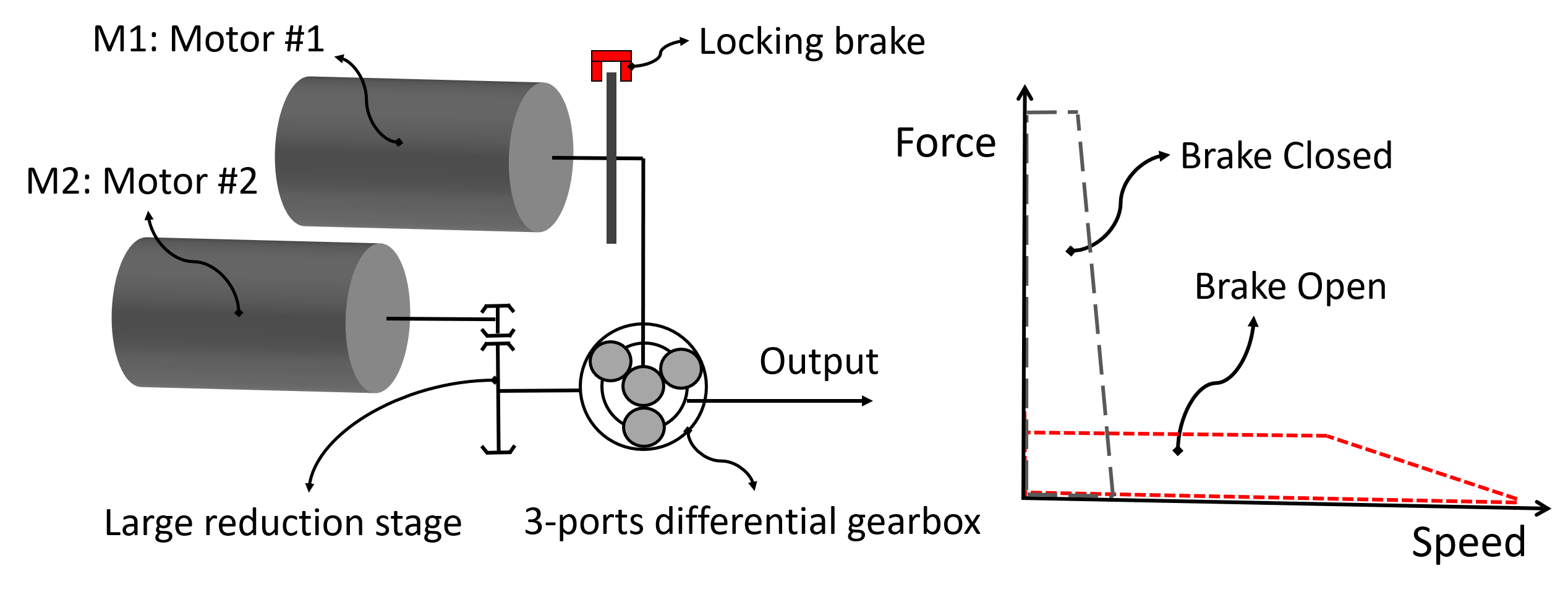}
		\vspace{-10pt}
	\caption{DSDM actuator architecture and force-speed curve}
	\label{fig:dualmotorconcept}
	\vspace{-10pt}
\end{figure}

The proposed architecture, referred to as a Dual-Speed Dual-Motor (DSDM) actuator, consists of an electric motor with a small reduction ratio (M1) equipped with a locking brake and another geared EM motor (M2) with a much larger reduction ratio coupled to the same output through a differential (see Fig. \ref{fig:dualmotorconcept}). The differential can be viewed as a junction where the speeds add up and the force is shared. 
The DSDM can be used in two modes: high-force (HF) mode when the brake is closed and high-speed (HS) mode when the brake is open. The result is like having two very different reduction ratios you can choose from during operation. The HF mode region is determined by the performance of M2 alone, since M1 is locked. The HS mode region can exceed the performance of M1 alone, since M2 can be used simultaneously to increase the output speed (motor speed adds-up see eq.\eqref{eq:kinematic}), increase the output torque (through inertial coupling see eq.\eqref{eq:f2}) and improve efficiency through an optimal distribution of M1 and M2 contributions. 
%
%

\subsection{Prototype design}
\label{sec:proto}

\begin{figure}[t]
	\centering
		\includegraphics[width=0.35\textwidth]{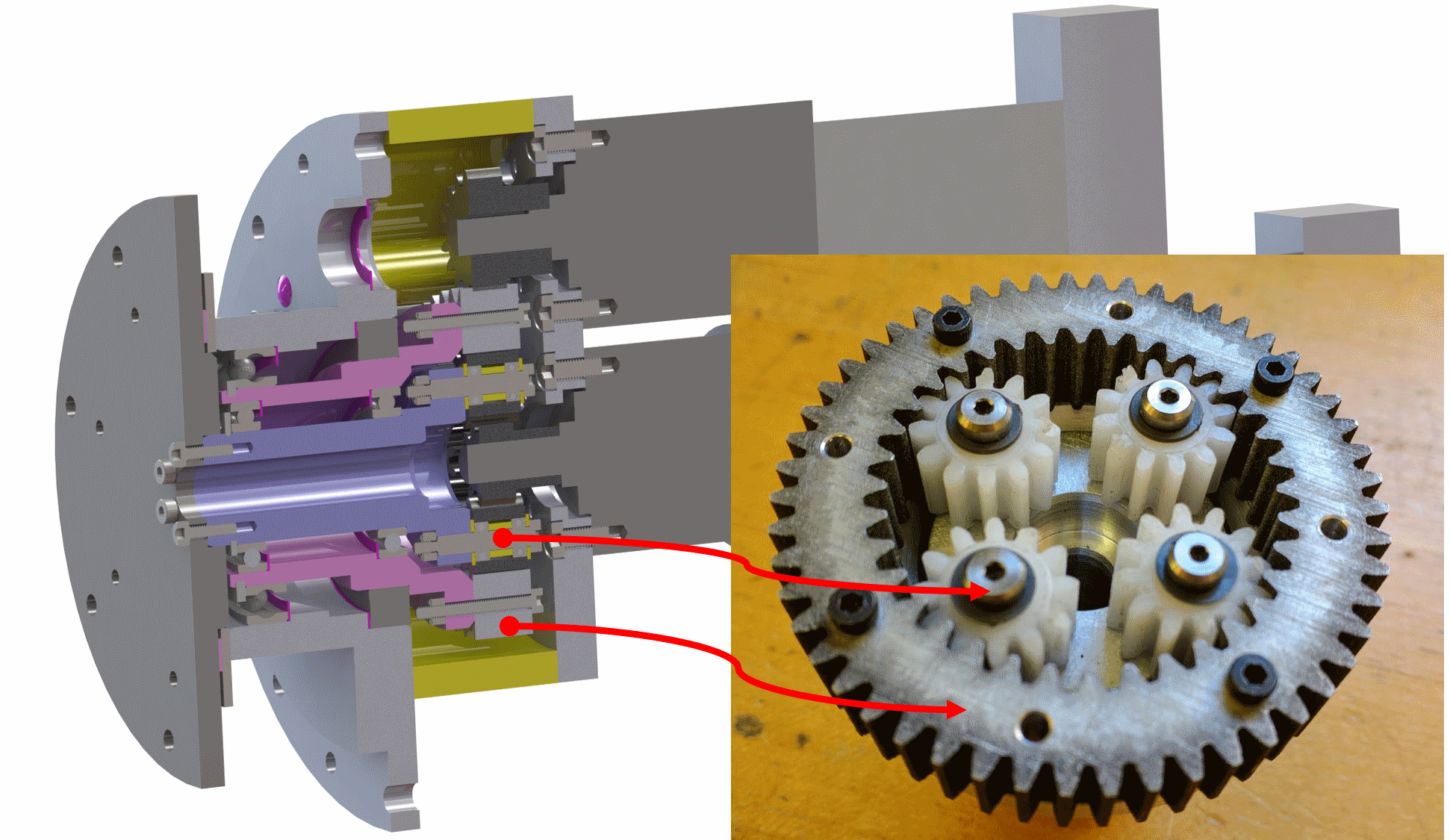}
	\caption{Section view of the CAD model of the DSDM actuator prototype} 
	\label{fig:dsdm_section}
\end{figure}

Shown in Fig. \ref{fig:dsdm_proto}, a prototype of a DSDM actuator embedded in a revolute joint was designed and built. The 3-port differential gearing was implemented with a custom-built planetary gearing, see Fig. \ref{fig:dsdm_section}. The planetary and output plate share the same bearings to save space. Two discrete \textit{Maxon} motors are used for the modularity and flexibility, a 90 watts motor for M1 and a 20 watts motor for M2. The gearing ratios were selected for the prototype to be backdrivable during HS mode, and to have an order of magnitude more torque during HF mode. The selected overall ratios and resulting specifications are given at Table \ref{tab:specrev}. The prototype weight approximately 1.5 Kg, about half of this weight is due to \textit{Maxon} motor assemblies (motors, gear-heads and brakes) and the other half is the custom built transmission and bearing assembly to support the output joint. Compared to a single-gear-ratio actuator of similar maximum torque and speed, i.e. a motor of about 350 watts with a 1:23 reduction, the DSDM prototype has many advantages. \textbf{1) Lightweightness}: Assuming the same level of mass optimization and same the power-density for the motors, the singe-gear actuator would be about 3 times heavier then the DSDM. (More details on the mass advantage of DSDM in \cite{girard_two-speed_2015}) \textbf{2) Efficiency}: The DSDM can hold a load with no power consumption in HF mode because of the large non-backdrivable 1:474 reduction, while the single-gear motor would need continuous current in this situation. Moreover, during motion, efficiency can be improved by selecting the motor in its most efficient conditions to provide power \cite{verstraten_modeling_2018} \textbf{3) Reliability}: An additional secondary advantage of the DSDM architecture is that some minimum performance can be guaranteed even if either motor fails because of the redundant motors connected to the same load.
\begin{table}[tb]
	\centering
	\caption{Specifications of the DSDM prototype}
		\begin{tabular}{ c c c c c }
			\hline
       Mode & Ratio & Max. Torque & Max. Velocity & Inertia \\ \hline
			      &       & Nm & RPM & kg m$^2$ \\
			\hline \hline
			HS & 1:23  &  2 & 220 & 0.004 \\			\hline
			HF & 1:474 & 14 &  10 & 0.22 \\			\hline
		\end{tabular}
	\label{tab:specrev}
\end{table}

\section{Modeling}
\label{sec:mod}

\begin{figure}[htbp]
 \vspace{-5pt}
	\centering
		\includegraphics[width=0.28\textwidth]{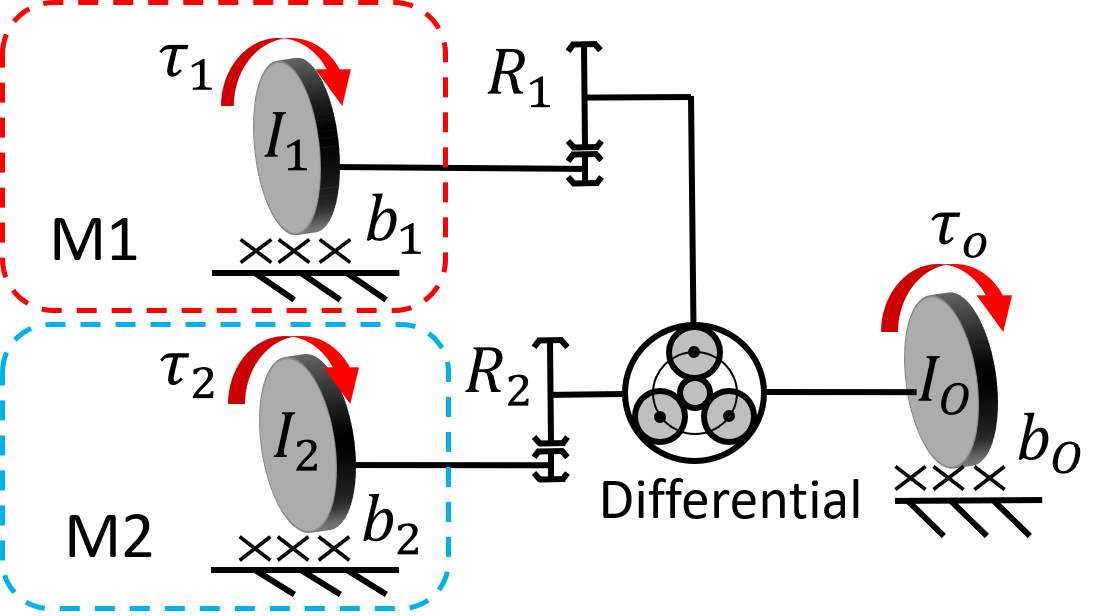}
		\vspace{-5pt}
	\caption{Lumped-parameter model of the DSDM}
	\vspace{-5pt}
	\label{fig:DSDM_model}
\end{figure}

A lumped-parameter model is proposed for the DSDM, illustrated in Fig. \ref{fig:DSDM_model}. The kinematic relationship is:
\begin{align}
\frac{w_1}{R_1 } + \frac{w_2}{R_2 } = w_o
\label{eq:kinematic}
\end{align}
where $w_1$, $w_2$, and $w_o$ are the angular velocity of M1, M2, and the output, respectively. Moreover, $R_1$ and $R_2$ represent the total overall reduction ratios and includes both the reduction thought the gear-heads and the planetary differential. The static force relationship is thus:
\begin{align}
R_1 \tau_1 = R_2 \tau_2 = -\tau_o
\label{eq:static}
\end{align}
where $\tau_1$ and $\tau_2$ are the electromagnetic torques of M1 and M2, and $\tau_o$ represents output-side external torques. When the brake is closed, $\tau_2$ include the brake reaction force.

Next, for the proposed dynamic model, the fast dynamics of the power-electronics is neglected and it is assumed that the inputs of the system are directly the electromagnetic torques $\tau_1$ and $\tau_2$. Also, all compliance of the components (gears and shafts) is neglected. Each port M1 ($i=1$), M2 ($i=2$) and output ($i=o$) is considered to have mechanical inertia and viscous damping represented by parameters $I_i$ and $b_i$. The system has 2 degree of freedom (DoF) when the brake is open, and 1 DoF when the brake is closed. The vector $\vec{w} = [w_o \; w_1]^T$ is chosen to describe the dynamic state of the system. 

\subsection{Discrete operating modes and transitions}

\begin{figure}[htbp]
 \vspace{-7pt}
	\centering
		\includegraphics[width=0.47\textwidth]{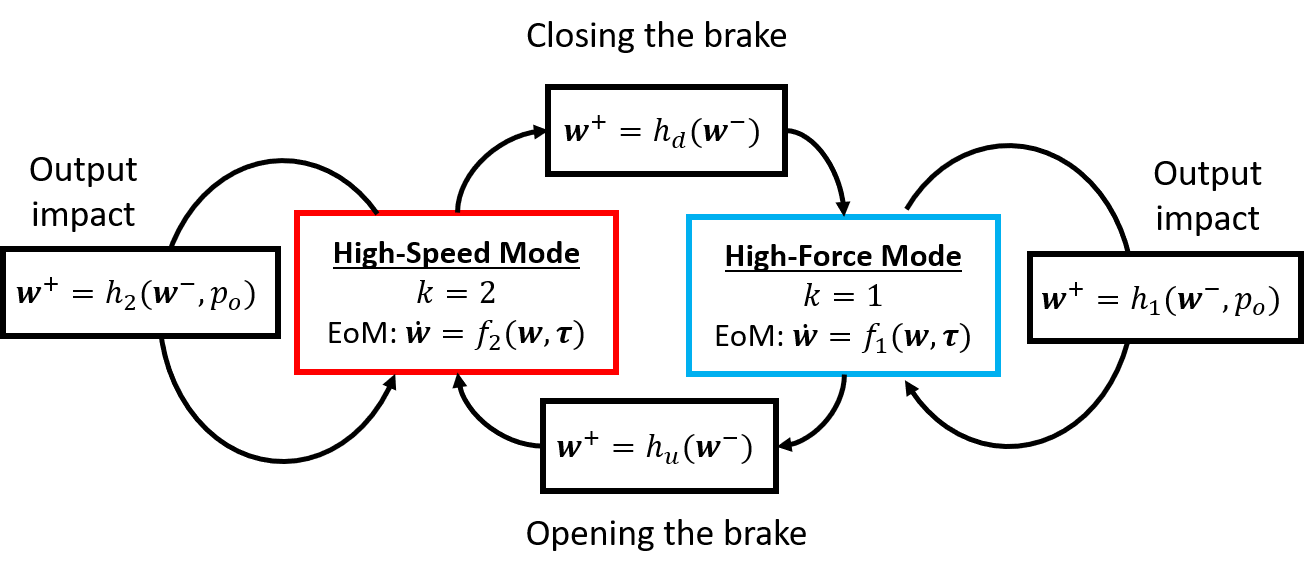}
		\vspace{-10pt}
	\caption{Discrete operating modes and possible transitions}
	\vspace{-5pt}
	\label{fig:operating_modes}
\end{figure}

Fig. \ref{fig:operating_modes} illustrates the two operating modes and all possible transitions. The DSDM has two discrete operating modes, each described by a differential equation: $\dot{\vec{w}} = f_k(\vec{w})$ with $k=1$ for HS mode when the brake is open and $k=2$ for HF mode when the brake is engaged. Also, transitions between these modes can lead to jumps of the dynamic states (impulsive behaviors) which are described by the mapping $\vec{w}^+ = h_u(\vec{w}^-)$ for an up-shift ($k:1\rightarrow2$) and $\vec{w}^+ = h_d(\vec{w}^-)$ for a down-shift ($k:2\rightarrow1$). Moreover, to describe impulsive behaviors due to contacts with the environment on the output side of the actuator, two additional jump maps are used: $\vec{w}^+ = h_2(\vec{w}^-,p_o)$ if the brake is open ($k=2$) and $\vec{w}^+ = h_1(\vec{w}^-,p_o)$ if the brake is closed ($k=1$), where $p_o$ is an impulse that will be discussed in section \ref{sec:impacts}.

When the brake is engaged, M1 is locked in place and the DSDM behaves like a regular geared-motor, where only the motor M2 drives the output through reduction ratio $R_2$. Equations of motion during this HF-mode are:
\begin{align}
\dot{\vec{w}} = f_1( \vec{w} , \vec{\tau} )&=
\left[ \begin{array}{c}
\scriptstyle \frac{1}{I_0 + R^2_2 I_2}  \left( -(b_o + R_2^2 b_2) w_o + R_2 \tau_2 + \tau_{o} \right)\\
0
\end{array} \right]
\label{eq:f1}
\end{align}
When the brake is open, both motor torque inputs contribute to the dynamic of the two states of the system. Equations of motion during this HS-mode are:
\begin{align}
\dot{\vec{w}} &= f_2( \vec{w} , \vec{\tau} ) = 
\left[ -H^{-1} D \right]
\vec{w} 
+ 
\left[ H^{-1} B \right] 
\vec{\tau}
\label{eq:f2}
\end{align}
where
\begin{align}
H &= 
\left[
\begin{array}{c c}
 \scriptstyle I_o + R_2^2 I_2           & \scriptstyle - \frac{R_2^2}{R_1} I_2       \\
 \scriptstyle - \frac{R_2^2}{R_1} I_2   & \scriptstyle I_1 + (\frac{R_2}{R_1})^2 I_2  \\
\end{array}
\right] \\
D &= 
\left[
\begin{array}{c c}
 \scriptstyle b_o + R_2^2 b_2           & \scriptstyle - \frac{R_2^2}{R_1} b_2       \\
 \scriptstyle- \frac{R_2^2}{R_1} b_2   & \scriptstyle b_1 + (\frac{R_2}{R_1})^2 b_2  \\
\end{array}
\right]\\
B &=
\left[
\begin{array}{c c c}
 \scriptstyle 0 & \scriptstyle R_2              & \scriptstyle 1 \\
 \scriptstyle 1 & \scriptstyle -\frac{R_2}{R_1} & \scriptstyle 0
\end{array}
\right] \quad 
\vec{w} = 
\left[
\begin{array}{l}
 w_{o} \\
 w_{1} 
\end{array}
\right] \quad
\vec{\tau} = 
\left[
\begin{array}{l}
 \tau_{1} \\
 \tau_{2} \\
 \tau_{o}
\end{array}
\right] 
\end{align}
\subsubsection{Up-Shift $k:1\rightarrow2$}
During high-force mode, the only DoF is described by the variable $w_o$. Opening the brake releases a constraint in the system and does not lead to any impulsive behavior. M1 rotor is simply suddenly free to move starting from rest. The mapping is thus given by:
\begin{align} 
\left[
\begin{array}{c}
w_o^+ \\ w_1^+
\end{array}
\right] = h_u( \vec{w}^- ) = 
\left[
\begin{array}{c}
w_o^- \\ 0
\end{array}
\right]
\label{eq:upshiftmap}
\end{align}
\subsubsection{Down-shift $k:2\rightarrow1$}
For a down-shift, the system goes from 2-DoF to 1-DoF, hence the sudden addition of a constraint (brake locked) can lead to an impulsive behavior. If the brake is engaged with non-zero $w_1$ velocity, the output will exhibit an undesirable impulsive deceleration:
\begin{align} 
\left[
\begin{array}{c}
w_o^+ \\ w_1^+
\end{array}
\right]
 = h_d( \vec{w}^- ) = 
\left[
\begin{array}{c}
w_o^- - \frac{w_1^-}{ R_1 \left( \frac{I_o}{I_2 R_2^2} + 1 \right) } \\ 0
\end{array}
\right]
\label{eq:downshiftmap_nonideal}
\end{align}
However, if M1 is at zero velocity before engaging the brake, then no impact will happen and the mapping is smooth:
\begin{align} 
\left[
\begin{array}{c}
w_o^+ \\ w_1^+
\end{array}
\right]
 = h_d( \vec{w}^- ) = 
\left[
\begin{array}{c}
w_o^- \\ 0
\end{array}
\right] \quad \text{if} \quad w_1^-=0
\label{eq:downshiftmap_ideal}
\end{align}
\subsection{Contact and external impacts}
\label{sec:impacts}
If a DSDM actuator is used on a robotic system making contact with objects, its internal velocities $\vec{w}$ can suddenly change due to impulsive forces. Impulsive output-side forces applied on the actuator output will be represented by $p_{o}=\int_{t^-}^{t^+} \tau_{o} dt$.
%
During high-speed mode, the impulsive map is:
\begin{align}
\left[
\begin{array}{c}
w_0^+ \\
w_1^+ \\
\end{array}
\right] = h_2( \vec{w}^-) = 
\left[
\begin{array}{c}
w_0^- \\
w_1^- \\
\end{array}
\right] +
H^{-1} 
\left[
\begin{array}{c}
p_o \\
0 \\
\end{array}
\right] 
\label{eq:h2}
\end{align}
If the reflected inertia of M2 is much greater than that of M1 ($R_2^2 I_2 >> R_1^2 I_1$), then the equation can be simplified to:
\begin{align}
\left[
\begin{array}{c}
w_0^+ \\
w_1^+ \\
\end{array}
\right] = 
\left[
\begin{array}{c}
w_0^- \\
w_1^- \\
\end{array}
\right] +
\frac{ p_{o} }{ I_o + I_1 R_1^2} 
\left[
\begin{array}{c}
1 \\
R_1 \\
\end{array}
\right]
\label{eq:hfimp}
\end{align}
Note that this simplification is adapted when the difference between the two ratios is large, like for the current prototype where $\frac{R_2^2 I_2}{R_1^2 I_1}\approx 425$.
Hence, during an impact, M2 velocity will be unchanged and the velocity discontinuity of the output will be transmitted directly to M1, from eq. \eqref{eq:hfimp} :
\begin{align}
\Delta w_1  =  R_1 \frac{ p_{o} }{ I_o + I_1 R_1^2} \quad                     
\Delta w_2  =  0                                         
\label{eq:dsdm_impact_gen_delta_w1}
\end{align}
During HF mode, assuming the brake is strong enough not to slip during the impact, the impulsive mapping is:
\begin{align}
\left[
\begin{array}{c}
w_0^+ \\
w_1^+ \\
\end{array}
\right] = h_1( \vec{w}^-) = 
\left[
\begin{array}{c}
w_0^- + \frac{ p_{o}}{I_o + R_2^2  I_2 } \\
0 \\
\end{array}
\right] 
\label{eq:h1}
\end{align}

\subsection{Nullspace of the DSDM during high-speed mode}
\label{sec:null}
%
%

\begin{figure}[H]
	\vspace{-2pt}
	\centering
		\includegraphics[width=0.30\textwidth]{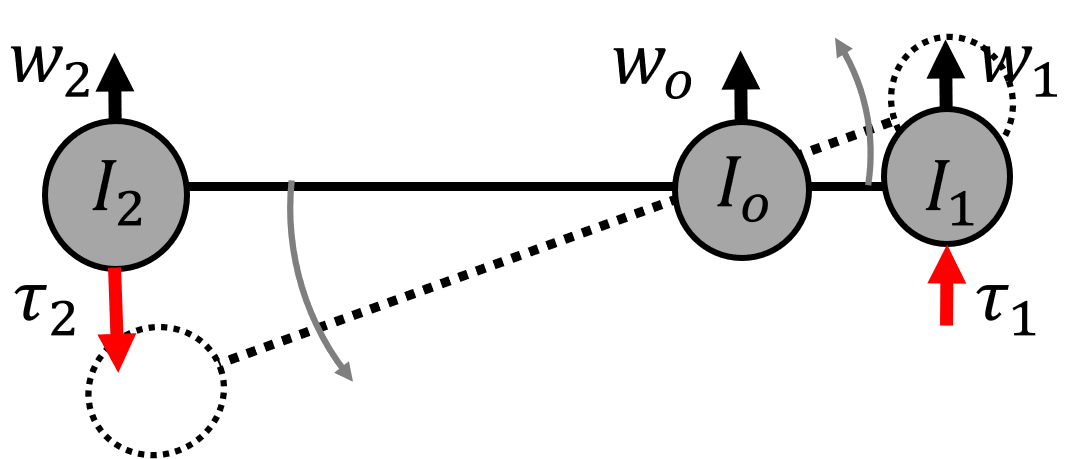}
	\vspace{-5pt}
	\caption{Nullspace illustrated graphically with a lever analogy}
	\label{fig:leverNullDyna}
	\vspace{-2pt}
\end{figure}

During high-speed mode, the DSDM actuator has one output and two torque inputs, hence there is one redundant DoF and a nullspace can be exploited. From the first line of eq.\eqref{eq:f2} and simplifying the expression by dropping the external force input and neglecting motor-side damping, the following input-output relationship is obtained:
\begin{align}
I_T \dot{w}_o +
b_T  w_o
=&
\left[ \begin{array}{c c}
R_1 & R_1 \frac{R_1 I_1}{R_2 I_2}
\end{array} \right]
\left[ \begin{array}{c}
\tau_1 \\
\tau_2
\end{array} \right]\\ 
 \scriptstyle I_T =    I_o + R_1^2 I_1 + \left(  \frac{R_1}{R_2} \right)^2 & \scriptstyle \frac{I_1}{I_2} I_o \quad\quad
 \scriptstyle b_T =   b_o + \left( \frac{R_1}{R_2} \right)^2 \frac{I_1}{I_2} b_o
\label{eq:output}
\end{align}
Hence, there is a 1-DoF space of inputs $\tau_1$ and $\tau_2$ that does not affect the output behavior:
\begin{align}
\left[ \begin{array}{c}
\tau_1 \\
\tau_2
\end{array} \right]
 = 
\underbrace{\left[ \begin{array}{c}
I_1 \\
-\frac{R_2 }{R_1 }  I_2
\end{array} \right]}_{\text{Nullspace Proj.}} u
\; \Rightarrow \; 
I_T \dot{w}_o +
b_T  w_o = 0 \quad \forall u
\label{eq:dyn_null_proj}
\end{align}
The nullspace can be interpreted graphically, see Fig. \ref{fig:leverNullDyna}, with an analogy where each port is represented by a mass connected to a lever representing the kinematic constraint. Eq. \eqref{eq:dyn_null_proj} can be interpreted as the ratio of forces at M1 and M2 ports that leads to a pure rotary motion about the output port. Interestingly the nullspace projection vector is only a function of intrinsic parameters, and independent of output-side parameters which are often unknown in practice. 
%

\section{Control Algorihtms}
\label{sec:ctl}

\begin{figure}[H]
\vspace{-2pt}
	\centering
		\includegraphics[width=0.45\textwidth]{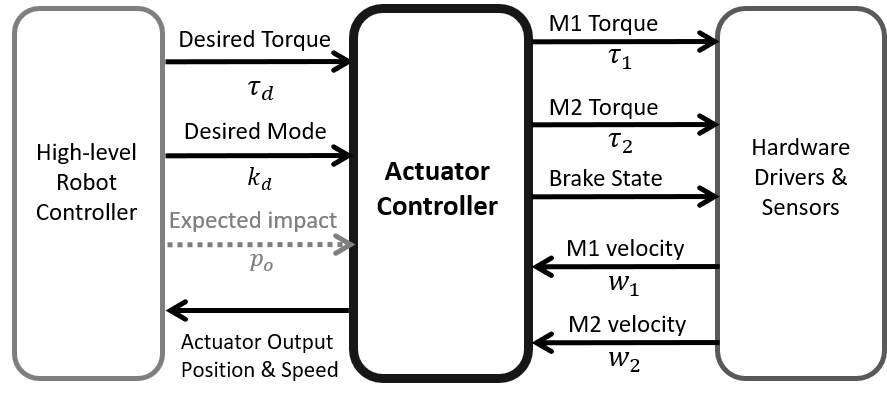}
		\vspace{-5pt}
	\caption{Control architecture of a DSDM controller}
	\label{fig:dsdm_control_architecture}
	\vspace{-5pt}
\end{figure}

Fig. \ref{fig:dsdm_control_architecture} shows the proposed hierarchical control architecture for a DSDM actuator. This paper discusses the actuator-level control laws coordinating motors during a shift process; the high-level robot controller was discussed in a previous publication \cite{girard_leveraging_2017}. The idea is to encapsulate DSDM actuators with simple control inputs: a desired motor torque $\tau_d$ and a desired discrete operating mode $k_d$, like a semi-automatic transmission in a car. The high-level controller, analogous to the driver using again a car analogy, then only must specify those two desired values ($\tau_d$, $k_d$), for each actuator of the robotic system, and is released from the low-level management of the gear-shift process. 

\subsection{Steady operation}

During steady operations, the commanded torque $\tau_d$ is directly sent to either M1 (HS mode) or M2 (HF mode):
\begin{align}
\left[ \begin{array}{c}
\tau_1 \\
\tau_2
\end{array} \right]
 =  
\left\{
\begin{array}{c@{}c}
\left[ \begin{array}{c} \tau_d \\ 0 \end{array} \right] & \text{ if } k=1  \\ \\
\left[ \begin{array}{c} 0 \\ \tau_d \end{array} \right] & \text{ if } k=2  \\
\end{array} 
\right. 
\label{eq:torquerooting}
\end{align}
This torque rooting scheme leads to simple input-output behavior. From eq.\eqref{eq:f1} and eq.\eqref{eq:f2} (with the assumption that $R_2^2 I_2 >> R_1^2 I_1$ and neglecting motor-side damping), the input-output dynamic is reduced to:
\begin{align}
\left[ I_o + R_k^2 I_k \right] \dot{w}_o +  \left[ b_o \right] w_o  &= \left[ R_k \right] \tau_d \quad k \in \{1,2\}
\label{eq:dsdm_output_R} 
\end{align}
Hence, from the point of view of the high-level robot controller, a DSDM actuator behaves like a single motor with two possible reduction ratio options, $R_1$ and $R_2$. 

\subsection{Transitions control}

\begin{figure}[h]
	\centering
		\includegraphics[width=0.47\textwidth]{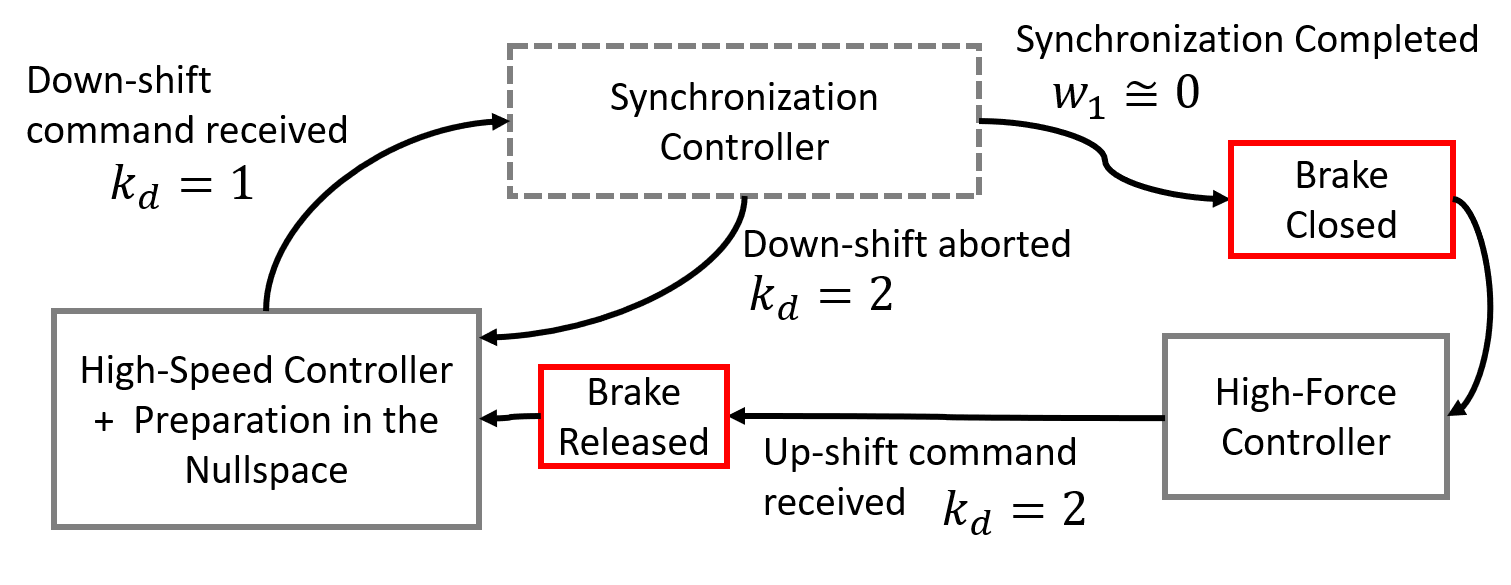}
		\vspace{-10pt}
	\caption{Operating modes of the DSDM actuator controller}
	\vspace{-5pt}
	\label{fig:opmodes}
\end{figure}

Fig. \ref{fig:opmodes} shows the operating modes of the proposed actuator-level controller and transition steps when a new operating mode command $k_d$ is received.

\subsubsection{Up-shift: $k:1\rightarrow2$}
For an up-shift, the transition is simple because the system goes from 1 DoF to 2 DoF. As described by eq. \eqref{eq:upshiftmap}, opening the brake at anytime does not lead to any undesirable impulsive behavior. As illustrated by Fig. \ref{fig:opmodes}, as soon as an up-shift is commanded the brake can be released; M1 is then instantaneously freed and the controller can immediately switch to the steady high-speed control mode.

\subsubsection{Down-shift: $k:2\rightarrow1$}
For a down-shift, the transition is harder because the system goes from 2 DoF to 1 DoF, and some synchronization work is needed. As described by eq. \eqref{eq:downshiftmap_nonideal}, closing the brake will lead to an undesirable impulsive behavior if M1 velocity is not zero. M1 speed $w_1$ must thus be brought to zero so that the locking brake can be engaged smoothly without any impact. Hence, as illustrated by Fig. \ref{fig:opmodes}, when a down-shift is commanded, an intermediary synchronization control mode is activated, and the brake is only closed after M1 has reached zero velocity. After the brake is engaged, the controller can then switch to the steady high-force control mode. 

\subsection{Synchronization and preparation controller}

When the brake is open (HS mode), it is proposed to leverage the nullspace with an internal speed controller. While the main loop continue to track the desired commanded torque $\tau_d$ (see eq. \eqref{eq:torquerooting}), a speed controller for $w_1$ can be projected on the output nullspace in parallel, with no influences on the output behavior:
\begin{align}
\left[ \begin{array}{c}
\tau_1 \\
\tau_2
\end{array} \right]
 = 
\underbrace{\left[ \begin{array}{c}
I_1 \\
-\frac{R_2 }{R_1} I_2 
\end{array} \right]}_{\text{Nullspace Projection}} \underbrace{C (w_{1,d} -  w_1)}_{\text{Speed controller}} + 
\underbrace{  \left[ \begin{array}{c}
\tau_d \\
0 
\end{array} \right]   }_{\text{Main Loop}}
\label{eq:prep}
\end{align}
This leads to an output dynamics unaffected by the speed controller (see Sec. \ref{sec:null}), and a $w_1$ closed-loop dynamic converging exponentially to a desired velocity $w_{1,d}$ if the synchronization gain $C$ is large. To accommodate torque and velocity saturations, the propose methodology is have \textbf{1)} a speed-controller with low gains always activated during HS operation (the preparation controller), and \textbf{2)} a speed-controller with high gains (the synchronization controller) when a down-shift command is received, to enforce aggressively $w_1 \rightarrow 0$ before engaging the brake. 
The preparation controller will aim at keeping $w_1$ as small as possible, by tracking $w_{1,d}=0$ during the regular operation, so that the synchronization will be faster when a down-shift command is received. Furthermore, when an impact is expected and can be estimated (for a legged robot while walking for instance), the preparation controller can take into account the incoming jump in velocity, and minimize the predicted M1 velocity right after the impact by tracking:
\begin{align}
w_{1,d}  = - R_1 \frac{ p_{o} }{I_o + I_1 R_1^2}
\end{align}
which will lead to $w_1^+=0$ if the tracking and impact prediction are perfect. In practice, $p_{o}$ estimation will be approximate, but using this scheme will help minimize the down-shifting time right after an impact.

\section{Experiments}
\label{sec:exp}

\begin{figure}[p]
 \vspace{-10pt}
	\centering
		\includegraphics[width=0.45\textwidth]{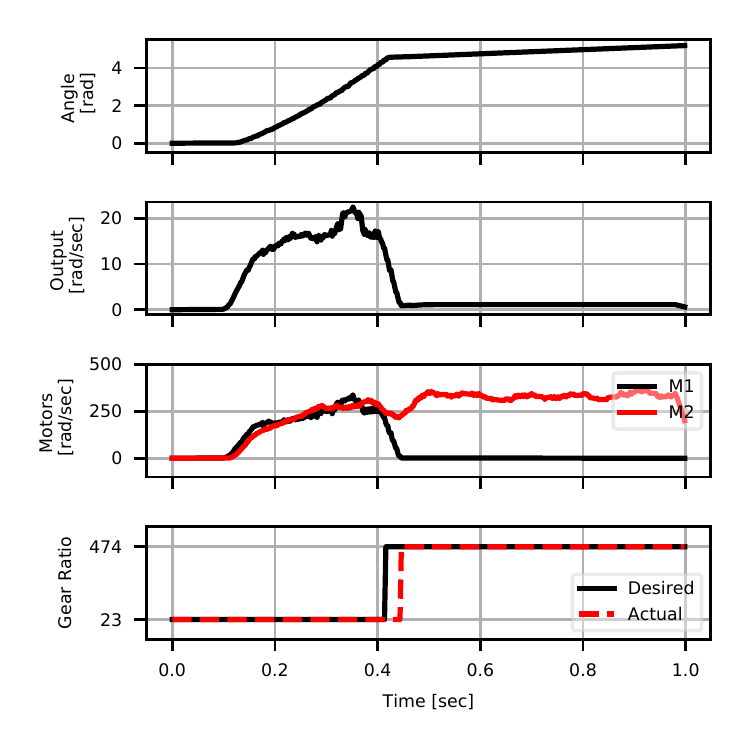}
		\vspace{-15pt}
	\caption{Down-shift during a contact with a stiff and heavy load}
	\vspace{-10pt}
	\label{fig:contact_exp}
\end{figure}

\begin{figure}[p]
 \vspace{-5pt}
	\centering
		\includegraphics[width=0.45\textwidth]{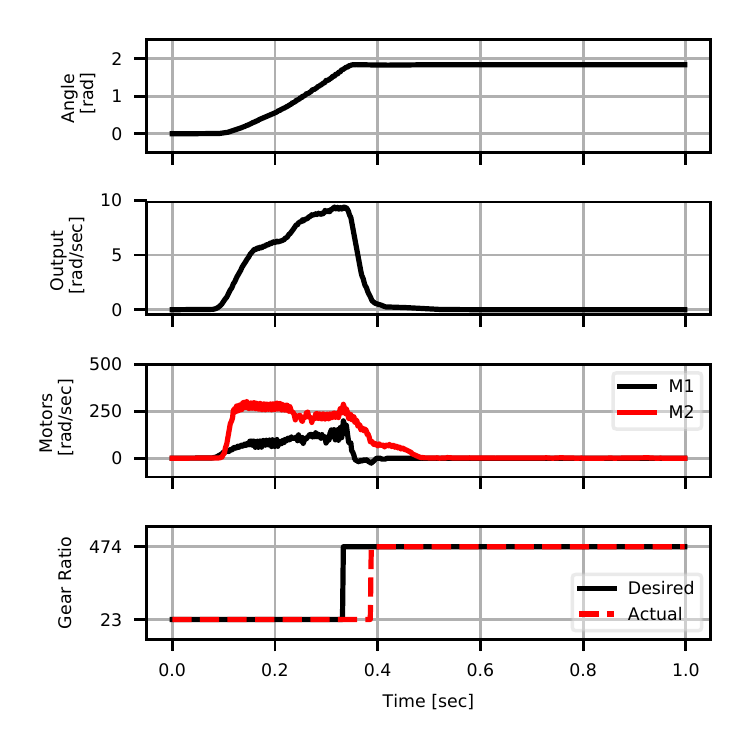}
		\vspace{-20pt}
	\caption{Down-shift during a contact with the ground (fixed object)}
	\vspace{-10pt}
	\label{fig:dsdm_fixed_down}
\end{figure}

\begin{figure}[p]
 \vspace{-5pt}
	\centering
		\includegraphics[width=0.45\textwidth]{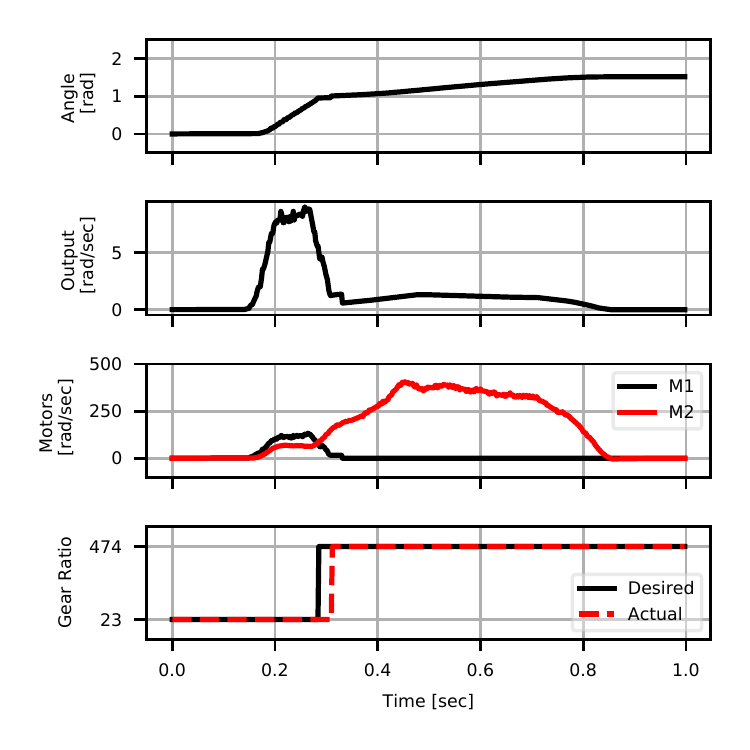}
		\vspace{-15pt}
	\caption{Down-shift during a contact with a compliant object}
	\vspace{-5pt}
	\label{fig:ballon_exp}
\end{figure}

The proposed gear-shifting methodology was implemented and tested with the prototype shown at Fig. \ref{fig:dsdm_proto}. Here three down-shift experiments during contacts are presented (also available in the video attachment). For all those experiments, the high-level loop is simply a constant torque $\tau_d$ command. Contact detection is done based on encoder measurements and triggers automatically the down-shift commands. Fig. \ref{fig:contact_exp} shows the DSDM making contact with a stiff heavy object (also corresponding to Fig. \ref{fig:dsdm_demo}). Results show that the DSDM actuator is able to engage high-force mode within 30 ms of the impact, and seamlessly continue its motion pushing the heavy object with large torques. Fig. \ref{fig:dsdm_fixed_down} shows the DSDM making contact with the ground, and also successfully engaging the HF mode very quickly. Finally, Fig. \ref{fig:ballon_exp} shows the DSDM making contact with a compliant load while also down-shifting successfully. Interestingly, down-shifts can be very fast in all those scenario representative of locomotion or manipulation tasks, since the reaction force of the contact actually helps the synchronization process by slowing down M1. This it not the case of most variable transmission architectures, many design are not adapted for shifting under load. Furthermore, the preparation controller make the DSDM use both motors during HS mode, and helps minimize the necessary synchronization time before engaging the brake when a down-shift command is received. 

%

\section{Conclusion}

In this paper, a two-speed actuator prototype capable of fast transitions between two operating modes that correspond to two effective reduction ratios, 1:23 or 1:474, is presented. A mathematical model of the hybrid behavior and control algorithms to coordinate the two motors of the actuator for fast transitions are proposed. Finally, experiments confirm that the proposed approach is capable of fast transitions during impact situations. The technology thus show great potential for tasks like locomotion and manipulation, that have typical operating sequences of 1) reaching at high-speed, 2) making contact and 3) bearing a large load.


%

\newpage
\newpage

\bibliographystyle{IEEEtran}
\bibliography{Zotero}

\begin{thebibliography}{10}
\providecommand{\url}[1]{#1}
\csname url@samestyle\endcsname
\providecommand{\newblock}{\relax}
\providecommand{\bibinfo}[2]{#2}
\providecommand{\BIBentrySTDinterwordspacing}{\spaceskip=0pt\relax}
\providecommand{\BIBentryALTinterwordstretchfactor}{4}
\providecommand{\BIBentryALTinterwordspacing}{\spaceskip=\fontdimen2\font plus
\BIBentryALTinterwordstretchfactor\fontdimen3\font minus
  \fontdimen4\font\relax}
\providecommand{\BIBforeignlanguage}[2]{{%
\expandafter\ifx\csname l@#1\endcsname\relax
\typeout{** WARNING: IEEEtran.bst: No hyphenation pattern has been}%
\typeout{** loaded for the language `#1'. Using the pattern for}%
\typeout{** the default language instead.}%
\else
\language=\csname l@#1\endcsname
\fi
#2}}
\providecommand{\BIBdecl}{\relax}
\BIBdecl

\bibitem{hollerbach_comparative_1992}
J.~Hollerbach, I.~Hunter, and J.~Ballantyne, ``A comparative analysis of
  actuator technologies for robotics,'' in \emph{The {Robotics} {Review}}, mit
  press~ed., 1992, vol.~2, pp. 299--342.

\bibitem{bologna_electric_2014}
S.~Bologna, ``Electric propulsion system for vehicles,'' U.S. Patent
  US8\,739\,655 B2, Jun., 2014, u.S. Patent.

\bibitem{lacerte_design_2016}
\BIBentryALTinterwordspacing
M.-O. Lacerte, G.~Pouliot, J.-S. Plante, and P.~Micheau,
  ``\BIBforeignlanguage{en}{Design and {Experimental} {Demonstration} of a
  {Seamless} {Automated} {Manual} {Transmission} using an {Eddy} {Current}
  {Torque} {Bypass} {Clutch} for {Electric} and {Hybrid} {Vehicles}},''
  \emph{\BIBforeignlanguage{en}{SAE International Journal of Alternative
  Powertrains}}, vol.~5, no.~1, pp. 13--22, May 2016. [Online]. Available:
  \url{http://saealtpow.saejournals.org/content/5/1/13}
\BIBentrySTDinterwordspacing

\bibitem{schoolcraft_gear_2011}
\BIBentryALTinterwordspacing
B.~Schoolcraft, ``Gear scheme for infinitely variable transmission,'' Patent,
  Aug., 2011, u.S. Patent. [Online]. Available:
  \url{https://patents.google.com/patent/US9228650B2/en}
\BIBentrySTDinterwordspacing

\bibitem{kembaum_ultra-compact_2017}
A.~S. Kembaum, M.~Kitchell, and M.~Crittenden, ``An ultra-compact infinitely
  variable transmission for robotics,'' in \emph{2017 {IEEE} {International}
  {Conference} on {Robotics} and {Automation} ({ICRA})}, May 2017, pp.
  1800--1807.

\bibitem{tahara_high-backdrivable_2011}
K.~Tahara, S.~Iwasa, S.~Naba, and M.~Yamamoto, ``High-backdrivable
  parallel-link manipulator with {Continuously} {Variable} {Transmission},'' in
  \emph{{IEEE}/{RSJ} {International} {Conference} on {Intelligent} {Robots} and
  {Systems} ({IROS})}, Sep. 2011, pp. 1843--1848.

\bibitem{hirose_design_1991}
S.~Hirose, K.~Yoneda, K.~Arai, and T.~Ibe, ``Design of prismatic quadruped
  walking vehicle {TITAN} {VI},'' in \emph{International {Conference} on
  {Advanced} {Robotics}}, Jun. 1991, pp. 723--728 vol.1.

\bibitem{jeong_design_2017}
S.~H. Jeong, Y.~J. Shin, and K.~S. Kim, ``Design and {Analysis} of the {Active}
  {Dual}-{Mode} {Twisting} {Actuation} {Mechanism},'' \emph{IEEE/ASME
  Transactions on Mechatronics}, vol.~22, no.~6, pp. 2790--2801, Dec. 2017.

\bibitem{hirose_development_1999}
S.~Hirose, C.~Tibbetts, and T.~Hagiwara, ``Development of {X}-screw: a
  load-sensitive actuator incorporating a variable transmission,'' in
  \emph{{IEEE} {International} {Conference} on {Robotics} and {Automation}
  ({ICRA})}, vol.~1, 1999, pp. 193--199 vol.1.

\bibitem{kim_improved_2007}
B.~Kim, J.~Park, and J.~Song, ``Improved manipulation efficiency using a
  serial-type dual actuator unit,'' in \emph{International {Conference} on
  {Control}, {Automation} and {Systems}}.\hskip 1em plus 0.5em minus
  0.4em\relax IEEE, 2007, pp. 30--35.

\bibitem{lee_new_2012}
H.~Lee and Y.~Choi, ``A {New} {Actuator} {System} {Using} {Dual}-{Motors} and a
  {Planetary} {Gear},'' \emph{IEEE/ASME Transactions on Mechatronics}, vol.~17,
  no.~1, pp. 192--197, 2012.

\bibitem{girard_two-speed_2015}
A.~Girard and H.~H. Asada, ``A two-speed actuator for robotics with fast
  seamless gear shifting,'' in \emph{{IEEE}/{RSJ} {International} {Conference}
  on {Intelligent} {Robots} and {Systems} ({IROS})}, Hamburg, Sep. 2015, pp.
  4704--4711.

\bibitem{verstraten_modeling_2018}
\BIBentryALTinterwordspacing
T.~Verstraten, R.~Furnémont, P.~López-García, D.~Rodriguez-Cianca, H.-L.
  Cao, B.~Vanderborght, and D.~Lefeber, ``Modeling and design of an
  energy-efficient dual-motor actuation unit with a planetary differential and
  holding brakes,'' \emph{Mechatronics}, vol.~49, pp. 134--148, Feb. 2018.
  [Online]. Available:
  \url{https://www.sciencedirect.com/science/article/pii/S0957415817301812}
\BIBentrySTDinterwordspacing

\bibitem{phlernjai_jam-free_2017}
\BIBentryALTinterwordspacing
M.~Phlernjai and T.~Takayama, ``Jam-free gear–clutch mechanism for
  load-sensitive step transmission in robotic joint,'' \emph{ROBOMECH Journal},
  vol.~4, p.~16, Jun. 2017. [Online]. Available:
  \url{https://doi.org/10.1186/s40648-017-0084-4}
\BIBentrySTDinterwordspacing

\bibitem{girard_leveraging_2017}
A.~Girard and H.~H. Asada, ``Leveraging {Natural} {Load} {Dynamics} with
  {Variable} {Gear}-ratio {Actuators},'' \emph{IEEE Robotics and Automation
  Letters}, vol.~2, no.~2, pp. 741--748, Apr. 2017.

\end{thebibliography}

\end{document}